\documentclass[11pt,a4paper]{article}
\usepackage{times,latexsym}
\usepackage{url}
\usepackage[T1]{fontenc}
\usepackage{graphicx}
\usepackage{amsmath}
\usepackage{comment}

\usepackage{subfigure}
\usepackage{parskip}

\usepackage[acceptedWithA]{tacl2018v2}

\usepackage{xspace,mfirstuc,tabulary}

\newif\iftaclinstructions
\taclinstructionsfalse 
\iftaclinstructions
\renewcommand{\anonsubtext}{(No author info supplied here, for consistency with
TACL-submission anonymization requirements)}
\newcommand{\instr}
\fi

\iftaclpubformat 

\else

\fi

\newcommand{\myparagraph}[1]{\vspace{2pt}\noindent{\textbf{#1.}}}
\newcommand{\red}[1]{\textcolor{red}{#1}}

\title{Visual Writing Prompts: \\Character-Grounded Story Generation with Curated Image Sequences}

\author{
  Xudong Hong$^{12}$, Asad Sayeed$^3$, Khushboo Mehra$^2$, \\
  {\bf Vera Demberg$^2$ and Bernt Schiele$^1$} \\[0.8ex]
  $^1$Dept. of Computer Vision and Machine Learning, MPI Informatics \\ 
  $^2$Dept. of Language Science and Technology, Saarland University \\ 
  $^3$Dept. of Philosophy, Linguistics, and Theory of Science, University of Gothenburg \\[0.8ex] 
  {\tt \{xhong,kmehra,vera\}@coli.uni-saarland.de}\\
  {\tt schiele@mpi-inf.mpg.de, asad.sayeed@gu.se}\\
}

\date{}

\begin{document}
\maketitle
\begin{abstract}

Current work on image-based story generation suffers from the fact that the existing image sequences collections do not have coherent plots behind them. 
We improve visual story generation by producing a new image-grounded dataset, Visual Writing Prompts (VWP).  
VWP contains almost 2K selected sequences of movie shots, each including 5-10 images. The image sequences are aligned with a total of 12K stories which were collected via crowdsourcing given the image sequences and a set of grounded characters from the corresponding image sequence. Our new image sequence collection and filtering process has allowed us to obtain stories that are more coherent and more diverse
compared to previous work. 
We also propose a character-based story generation model driven by coherence as a strong baseline. Evaluations show that our generated stories are more coherent, visually grounded, and more diverse than stories generated with the current state-of-the-art model. \red{This is a pre-MIT Press publication version. }


\end{abstract}

\section{Introduction}
\begin{figure*}[ht]
\includegraphics[width=0.95\textwidth,trim={0cm 5.5cm 0cm 0cm},clip]{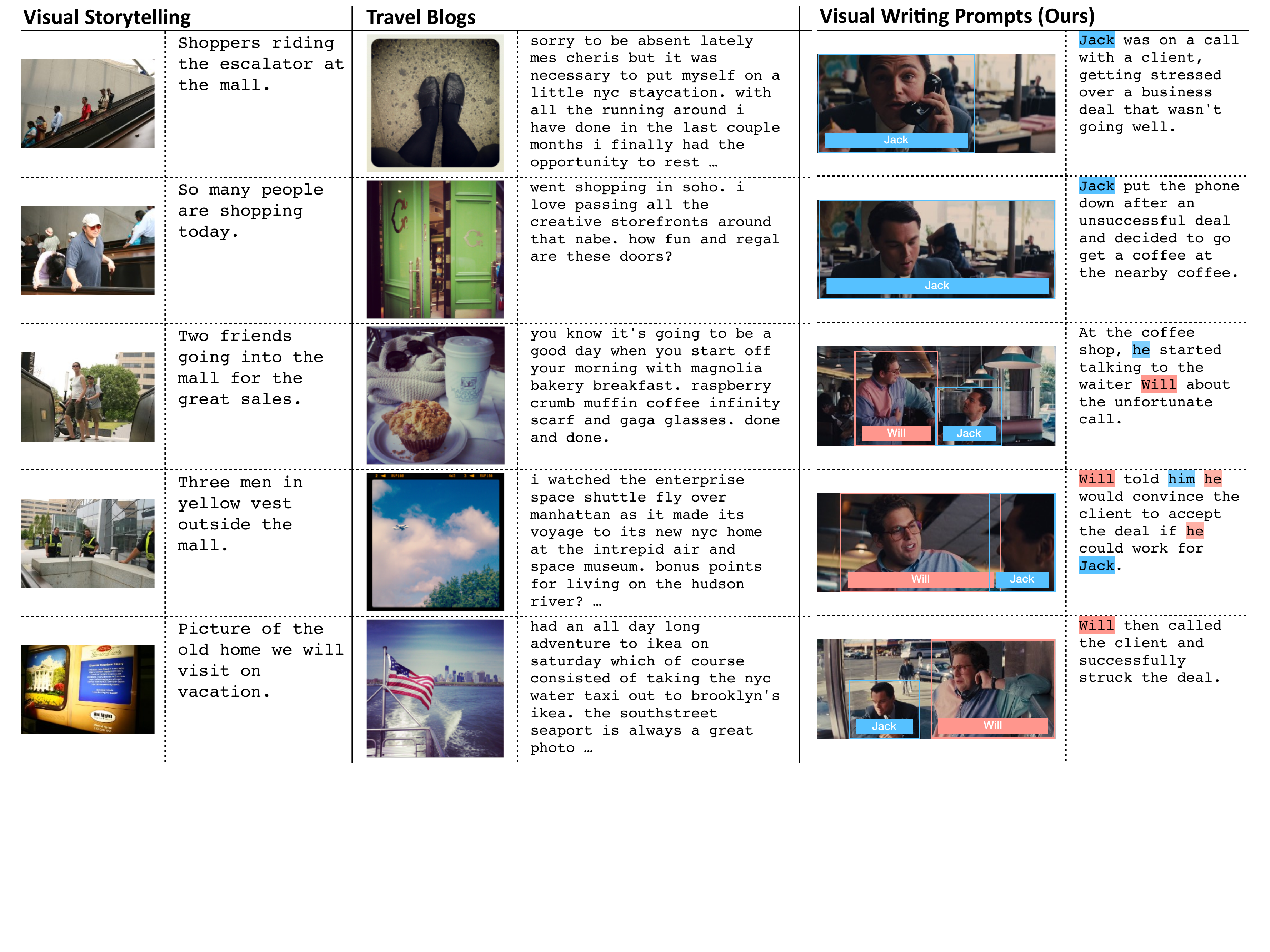}
\caption{Comparison of Visual Writing Prompts dataset with Visual Storytelling and Travel Blogs datasets. Our dataset has recurring characters across all five images and sub-stories. Each appearance of a character in a sub-story has a bounding box in the corresponding image, which grounds the textual appearance to visual input. }
\label{fig:datasets-comp}
\end{figure*}

In this work, we improve the quality of text stories generated by neural models from image sequences.  We do so by improving the curation of the image sequences that form the basis for collecting the story/image pairs used to train the models: we build a dataset in which the images lend themselves better to telling a story. 
To show the usefulness of our dataset, we train a coherence-driven model where we design a coherence component inspired by entity grid models. Experiments show that our model produces more coherent, visually-grounded stories with more diversity than previous models\footnote{We will release our code, image features, annotations and collected stories on a website. }. 

Stories play an important role in natural language understanding and generation because they are the key mechanism for humans to understand meaning and knowledge in the world \citep{piper-etal-2021-narrative}. Automatically generating a coherent and interesting story is a complex task requiring various capabilities in language processing, event comprehension, and world knowledge to come together. Previous approaches to story telling have used different kinds of input to guide the story: 
some use a textual prompt to start the story \citep{fan2018hierarchical}. 
Yet others involve describing a sequence of images to direct the story \citep{Huang2016}. We choose to work inside the latter family of approaches in order to exploit the rich information contained in characters and to prevent suffering from the grounding problem \citep{harnad1990symbol}. 

Research on visual narratives shows how it would be possible to construct the sort of dataset we propose: image sequences should consist of a series of coherent events centered around one or more main characters \citep{cohn2020visual}. In fact, even Aristotle already points out in \textit{Poetics} that \textit{event} and \textit{character} are the most important elements for a good story. 

To date, several datasets of image sequences for narrative generation exist, such as the Visual Storytelling  \citep[VIST;][]{Huang2016} dataset, which includes sets of images extracted from Flickr albums. However, image sequences generated this way have the drawback that they may not lend themselves well to storytelling. Consider for instance the image sequence shown in the first column of Fig. \ref{fig:datasets-comp}: the people featured across the image sequence are all different, and there is no real development of an event or a plot. This means that the stories that humans were able to write for these types of image sequences are often quite poor from a narrative point of view and therefore lead to low-quality training data for our story generation algorithms, which in turn, unsurprisingly, generate quite bad stories.




We thus argue that image sequences serving as writing prompts should be comprehensible as visual narratives by humans. 
Humans (with reasonable writing proficiency) can then ``translate'' such visual narratives into textual narratives. 
For an image sequence to qualify as a visual narrative, events and characters must have two properties: \textit{coherence}, meaning that the events are semantically related and centered around recurring characters; and \textit{diversity}, meaning that several different events jointly construct a plot. 
Psycholinguistic experiments show that missing either of these properties impedes human comprehension of image sequences as visual narratives \citep{cohn2012pea}. 
In addition, the characters should also be easily recognized in the image sequences and can be straightforwardly linked to the stories ({\it visual groundedness}). 
Image sequences without these properties are hence not effective writing prompts. 

In this work, we define the term \textit{visual tellability} to mean the \textit{tellability} \citep{huhn2014handbook} of image sequences, i.e.,~how likely it is that humans can write a story with an image sequence, which measures whether the image sequences have the two properties described above. We propose a new dataset, Visual Writing Prompts (\textbf{VWP}), containing curated image sequences and matching user-generated stories, linking the image sequences into coherent stories. Our image selection process allows us to choose optimized image sequences that have high visual tellability, and to encourage our crowdsourced storytellers to produce coherent stories with high diversity. 


To obtain coherent and visually grounded stories, we provide cropped images of characters explicitly with image sequences for storytellers. 
To improve narrativity and diversity, we select images from a data source that is already likely to have a plot: image sequences selected from movie scenes with aligned synopses. 
To further show the importance of coherence and visual groundedness, we propose a story generation model with a representation of visual coherence focused principally on character continuity as a strong baseline. Experiments show that our model outperforms the current state-of-the-art and generates stories that are more coherent, visually grounded, and have higher diversity. 



We summarize our contributions in this work as follows: 
\textbf{(a)} We propose a pipeline to extract images sequences automatically from annotated movies as story writing prompts, which leads to image sequences with higher visual tellability. 
\textbf{(b)} We collect a new dataset of stories based on curated image sequences with grounded characters, which is more coherent and has better diversity than previous datasets. 
\textbf{(c)} We propose a character-grounded story generation model driven by visual coherence as a strong baseline for image-based story generation, which generates more coherent, diverse and visually grounded stories than the current state-of-the-art model TAPM \citep{yu2021transitional}. 



\section{Related Work}
\myparagraph{Story generation} 
Several datasets have been presented for generating a story conditioned on a prompt such as title ~\citep{fan2018hierarchical}, keywords~\citep{yao2019plan}, cue phrases~\citep{xu-etal-2020-megatron} or story plot~\citep{rashkin2020plotmachines}. The ROCStories corpus ~\citep{mostafazadeh2016corpus} is a collection of short stories with rich causal and temporal relations. 
In subsequent work, new datasets have also been formed by gathering annotations on subsets of ROCStories for specialized story generation tasks such as modeling character psychology  \citep{rashkin2018modeling} counterfactual reasoning  \citep{qin2019counterfactual}, etc. 
The STORIUM dataset \citep{akoury2020storium} of collaboratively-written long stories contains rich annotations such as narrator prompts, character goals, and other attributes to guide story generation. 
However, all these datasets relying on textual prompts suffer from the grounding problem that the meanings of textual stories are grounded on textual symbols \citep{harnad1990symbol}. In contrast, our dataset contains stories grounded on characters in image sequences, i.e. nonsymbolic prompts from visual perception. 

\label{sec:vst}
\myparagraph{Visually-grounded stories} 
Early work on the VIST dataset \citep{Huang2016} identified that language in visually-grounded stories is much more diverse than in image captions. 
However, most of the previous datasets of visually-grounded stories have limitations because characters are not explicitly annotated \citep{chandu2019storyboarding}, the dataset is limited in scale \citep{xiong2019synopses}, or there is no sequence of events behind the images \citep{park2015disneyNYC, Huang2016}. 
Our dataset is the first large-scale dataset that is focused on overcoming these limitations. 
Unlike the VIST images, images in our VWP dataset do not feature people posing for the camera in limited contexts. Instead, they depict a rich range of situations, interactions, and emotions. Furthermore, providing character annotations in VWP ensures that the entities in the narrative are grounded to the image sequence and can be easily tracked across the sequence even when some visual attributes change. We hypothesize that these features will result in more coherent and visually grounded stories while maintaining a high level of diversity.  

\section{Image Sequence Construction}
\label{img_seq_construct}

 \begin{figure}[t]
\includegraphics[width=0.48\textwidth,trim={7cm 10cm 7cm 0cm},clip]{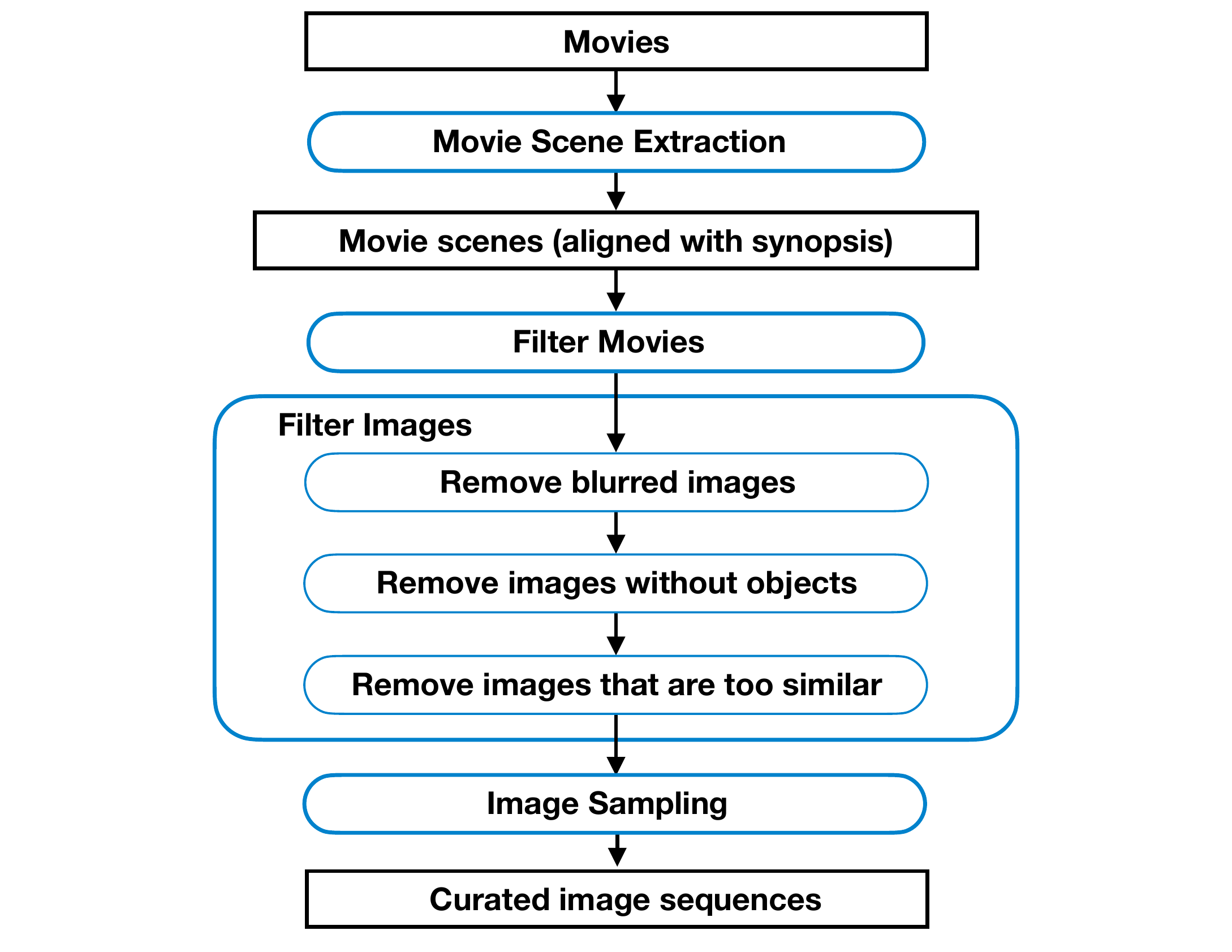}
\caption{Image processing pipeline. Black squares are input or output. Circles are processing steps. }
\label{fig:pipeline}
\end{figure}

\begin{figure*}[ht]
\centering
\includegraphics[width=0.8\textwidth]{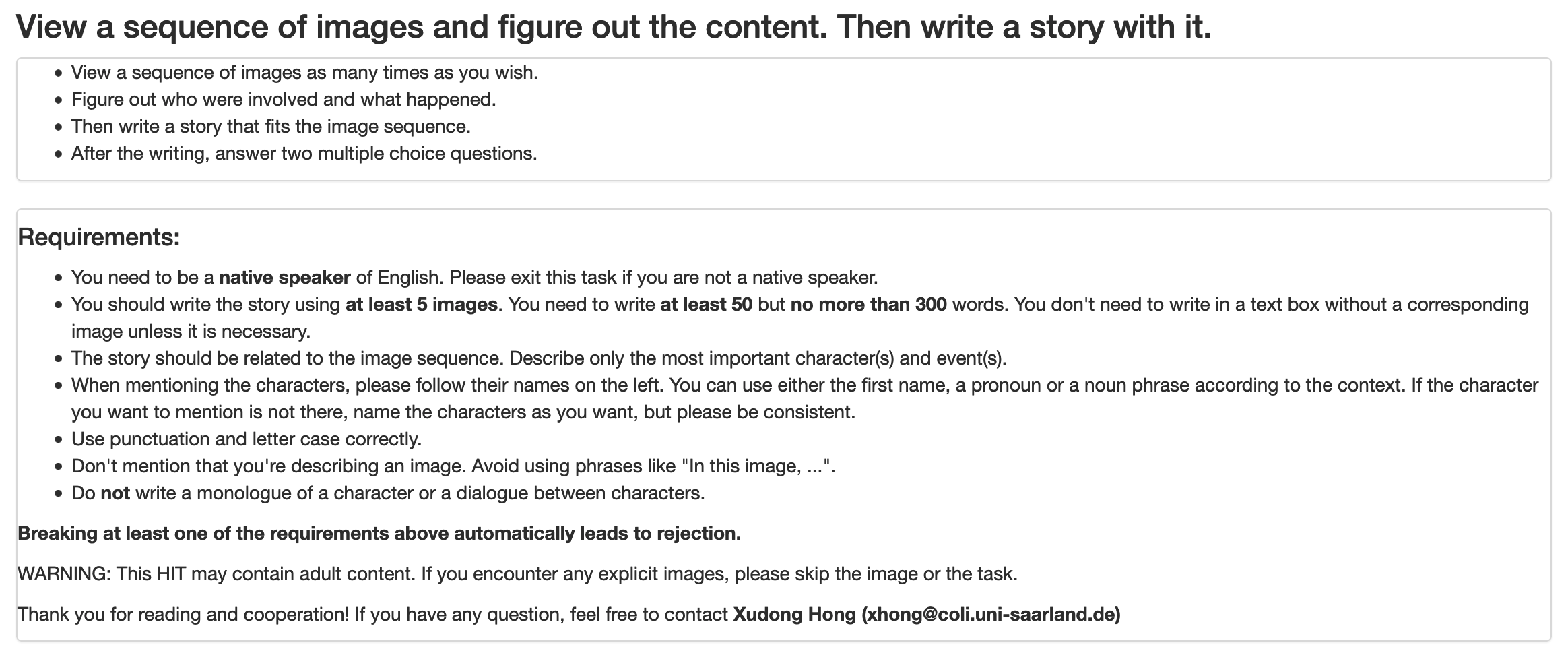}
\includegraphics[width=0.8\textwidth]{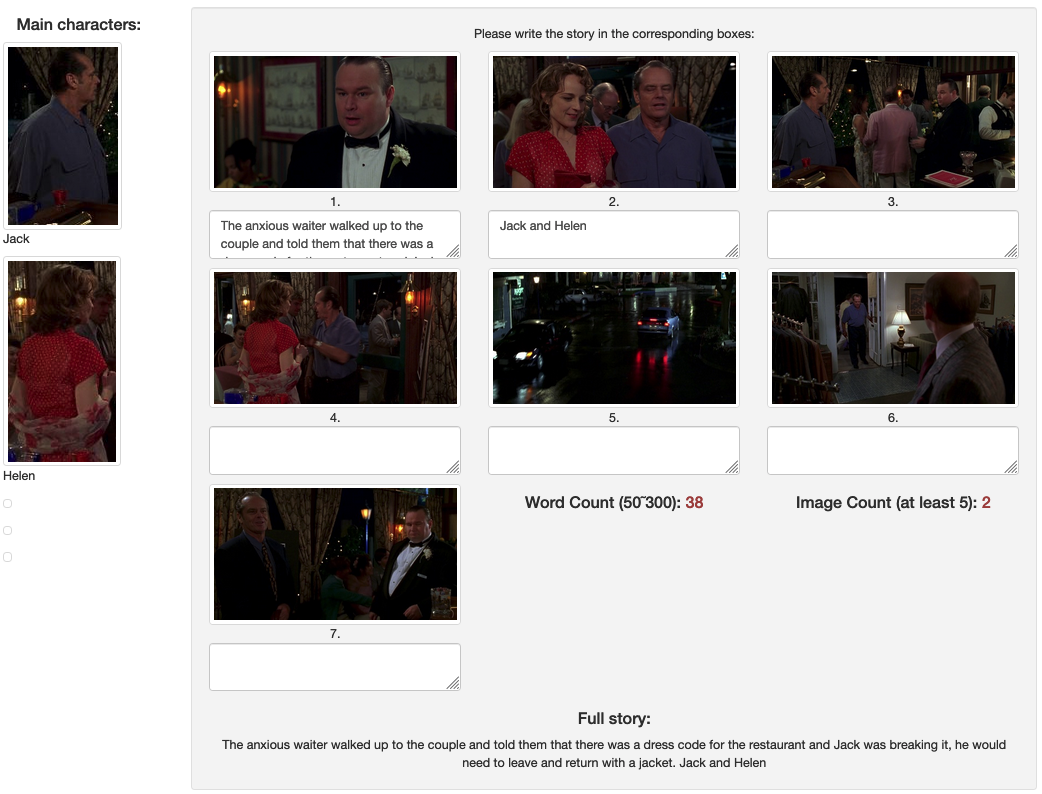}
\caption{Worker interface on Amazon Mechanical Turk. The instructions and the requirements are presented first. The main characters are provided on the left side. On the right side, each image is accompanied by a $textarea$. The full story is presented under the input area. We also show the word count and the number of images used for workers' convenience. }
\label{fig:amt-gui}
\end{figure*}

In this section, we describe how we obtain image sequences and design a pipeline to filter and sample images. 
Our objective is to construct image sequences that are \textit{visually tellable}, i.e. are coherent and have high diversity. 
Our full pipeline for image sequence construction is shown in Figure \ref{fig:pipeline}.

\myparagraph{Movie Scene Extraction} 
To achieve high coherence and diversity, we choose to select images from movie scenes that have a plot consisting of a series of events around several main characters. 
We extract movie scenes from MovieNet \citep{huang2020movienet} since it is a dataset that contains movie synopses, annotated movie scenes with extracted movie shots and identified main characters. The paragraphs in each movie synopsis describe sub-plots of the movie plot, which are aligned with one or more movie scenes.  

Changing from one paragraph to another in the synopsis indicates scene changes \citep{xiong2019synopses}. Moreover, events and characters in one movie scene are semantically coherent. We can make use of these properties to achieve high diversity by sampling image sequences from movie scenes aligned with only one paragraph, so that image sequences are from one sub-plot with a series of different events. 



\myparagraph{Filtering Movies} 
Since we want to constrain the range of commonsense inferences of storytellers to the real world and help them to produce coherent stories, we first filter out all fantasy, science fiction, and horror movies. We also filter out all animations because their image characteristics are too different from the other movies.

\myparagraph{Filtering Images}\footnote{Hyper-parameters in this section are determined by a preliminary experiment that optimizes the filter process manually on 50 image sequences.}
\label{img_seq_construct}
To help storytellers to write stories that are visually grounded on characters or objects around them, we discard blurry images and images without any COCO ``objects''
\citep{lin2014microsoft}\footnote{A human character is also labeled as an ``object'' in MSCOCO dataset for object detection. }. 
We measure the amount of image blur by calculating the variance of the Laplacian \citep{pech2000diatom} and remove images with a variance lower than $30$. 
We further apply a MaskRCNN-based object detector and filter out images without any detected objects -- this will help us generate stories with interesting grounding in the image. 

To increase the diversity of image sequences, we need to avoid including shots that are very similar (as can happen when a person speaks in a long monologue, for example) to one another. To detect the degree of similarity, we first feed the images to a ResNet-50 pre-trained on ImageNet and extract image features after the $fc7$ layer. Then we compute pairwise cosine similarities of the image features within each image sequence and discard an image if its cosine similarity with any one of the other images is larger than $0.89$. 

Additionally, we detect adult content by applying a pre-trained classifier\footnote{\url{https://github.com/notAI-tech/NudeNet/}} and exclude images that trigger the classifier. We also remove the first or the last image sequence in a movie to avoid images with credits.

\myparagraph{Image Sampling} 
The most intuitive way to collect stories is to use extracted movie scenes directly as writing prompts. Since these movie scenes contain a large number of movie shots, we control the workload by constraining the number of images for each writing task to a lower number $K$ which is obtained through the 2nd pilot studies in Section \ref{sec:pilot}. So from each selected movie scene, we sample images consecutively in non-overlapping sliding windows with a size of $K$ and use each set of $K$ images as one writing prompt.



\section{Crowdsourcing Experiment Design}

In this section, we design a crowdsourcing experiment to collect stories using our collected image sequences as writing prompts. 
Our objective is to obtain coherent stories that have high diversity from crowdsourced storytellers.

We design and run all of our studies on Amazon Mechanical Turk (AMT). The worker user interface is shown in Figure \ref{fig:amt-gui}. 
In each assignment, we ask the worker to select a subset of images from the image sequence and write a short story (50 to 300 words) that fits the image sequence. 
To ensure the human-written stories are grounded on main characters, we provide names and cropped images of at most five major characters. We retrieve the bounding boxes for each character from the MovieNet annotations and choose the least blurry appearance of each character in the image sequence. 
We pose three questions to the workers. 
The first two questions are used to identify workers who have watched the movie from which the image sequence is taken, as they might exhibit different behaviors during story-writing. The third question is to measure the visual tellability on a 5-point Likert scale, which is used to show the effectiveness of our image sequence construction pipeline. 

We also design a review form for story reviewers to judge the quality of collected stories. We ask the reviewers: 1) whether they want to approve the story; 2) if not, which requirement does it break? 3) if yes, judge the statement: \textit{this is a good story.} on a 5-point Likert scale. 
The first two questions are to assure that the collected stories fulfill the requirements including: the story is grammatical, the story has high diversity and the story is visually grounded. 
The third question is to get judgments of the quality of the approved stories.


\subsection{Pilot studies}
\label{sec:pilot}
We identify the following design questions of the crowdsourcing experiment for data collection: 

\myparagraph{1} Does the image filtering process improve the tellability of the image sequences? 

\myparagraph{2} What is the optimal number of images to provide to workers to achieve high visual tellability at a reasonable workload in one writing prompt?




\noindent We conducted two pilot studies to investigate these questions. 
We collect 5 stories per image sequence at most from different writers. 
 

\myparagraph{Pilot study 1: Effectiveness of image filtering} 
The first study tests whether our image-filtering steps (see Section \ref{img_seq_construct}) increase the visual tellability of the extracted image sequences.
We extract 180 movie scenes containing 10 images each from selected movies; on half of these, we apply our image filters, while we leave the others as is. All resulting image sequences have 5 to 10 images.  

Results show that the average visual tellability score of image sequences with filtering is 3.7, which is significantly higher (unpaired $t$-test, $t=4.89$, $p$-value $<0.001$) than the average visual tellability score of image sequences without filtering (3.29). This shows that our image filtering process in the image sequence construction pipeline leads to higher visual tellability and we will apply image filtering in our data collection. 

\myparagraph{Pilot study 2: Number of images to display} 
The second study explores the effect of the number of images $K$ in a writing prompt on workload and visual tellability. We randomly sample 150 movie scenes with 20 images, where writers can choose from 5 to 20 images for their stories. We set the minimum number of images to 5 because the most common narrative structure is \textit{5-part play} that contains five components \citep{cohn2013visual}. In addition, since there are five sentences per story in both ROCStories and VIST datasets, we can make the stories with 5 images in our dataset comparable to theirs. 
We set the maximum number to 20 because we find in a preliminary experiment that the workload of writing prompts with more than 20 images is too high considering our budget. We then run our study on these scenes. 

We find a negative correlation between the actual number of images used by storytellers and the visual tellability scores, $r(500)=-0.17$, $p<0.001$. This result indicates that showing fewer images can both improve visual tellability and reduce workload. 
However, we also want to obtain longer stories, we prefer to have a $K$ larger than 5. 
Since a majority of $89\%$ of the human-written stories use 5 to 10 images out of 20 and achieve a reasonably high average visual tellability (3.75), we set the maximum number of images we display to 10. 

\begin{table*}[t]
\centering
\begin{tabular}{l|llllll}

\textbf{Name}	&	\textbf{Image}	&	\textbf{\# Text}	&	\textbf{\# Image}	&	\textbf{\# token}	&	\textbf{\# Event}	&	\textbf{\# Char.}	\\	
	&	\textbf{Genre}	&		&	\textbf{per Text}	&	\textbf{per Text}	&	\textbf{per Text}	&	\textbf{per Text}	\\	\hline
VIST	&	photos	&	50 K	&	5	&	57.6	&	6.3	&	3.4	\\	
Travel blogs	&	photos	&	10 K	&	1	&	222.3\ddag	&	3.8\ddag	&	2.3\ddag	\\	
VWP (Ours)	&	movie shots	&	12 K	&	[5, 10]	&	83.7	&	12.8	&	13.1	\\	\hline					

\end{tabular}
\caption{\label{tab:dataset-statistics} Comparison of statistics of VWP against previous datasets. Numbers with \ddag~are obtained from a small sample of the Disney split of the dataset that is available in their repository. }
\end{table*}

\begin{table}[t]								
\centering								
								
\begin{tabular}{l|l|ll}								
\textbf{Dataset}	&	\textbf{\# stories}	&	\textbf{LL}	&	\textbf{Avg. LL}	\\	\hline
VIST	&	4987	&	-4017	&	-0.8055	\\	
VWP (Ours)	&	4680	&	\textbf{-3722}*	&	\textbf{-0.7953}*	\\	\hline
\end{tabular}														

\caption{\label{tab:local-coherence} Coherence by log-likelihood (LL) and average log-likelihood (Avg. LL) on validation split of VIST versus a sample split from our VWP dataset with the same number of image sequences. The stories are more coherent if the number is larger.  } 
\end{table}

\section{Data Collection}

In this section, we describe how we collect and process the stories in the VWP dataset. 
Our goal is to obtain narratives given the curated image sequences as writing prompts. 


\myparagraph{Worker Qualification} In order to improve story quality, we apply a qualification process to workers. We first collect 4.5K stories together with visual tellability judgments and obtain 556 candidate workers. Each story is reviewed by one of five graduate students. To ensure that the reviewers mutually understand the purpose of the task, we let the reviewers judge 100 stories then check the reviews together to agree on the judgment standards. 
We then select 58 qualified workers with an acceptance rate $\geq$90\%, average story quality $>$3.1, and accepted assignments $\geq$5. We assign a qualification to these workers and invite them to bulk collection.

\myparagraph{Bulk Collection} 
We collect 7.5K stories with the qualified workers in bulk collection. We group about 300 image sequences into a batch and collect 1.5K stories per batch. For each batch, we sample $s$ stories from each worker and review the stories to update the assessment of the worker,  
$$ s =\left\{
\begin{aligned}
& 10, & \text{ if } n_w <10 \\
& 10 \log n_w, & \text{ otherwise} \\
\end{aligned}
\right.
$$
where $n_w$ is the number of stories that worker $w$ wrote in this batch. 
We run the bulk Collection batch by batch and revoke the qualification if the worker does not satisfy the selection criteria anymore.

\myparagraph{Text Processing} 
We process the raw text to make it easier for training story generation models. 
We tokenize all stories with the spaCy\footnote{\url{https://spacy.io/}} English tokenizer. We then recognize all entities using a Name Entity Recognition model \cite{Peters2017SemisupervisedST}. We 
change all location names to placeholders and replace all named characters in each story to $[male0], ..., [maleM], [female0], ..., [femaleN]$. We obtain the gender of each named person based on a name statistics\footnote{\url{https://ssa.gov/oact/babynames/names.zip}}. Finally, to mark the alignment between images and story sections, we add a special separator token $[sent]$. We randomly sample 849 stories as \textit{validation} split and 586 stories as \textit{test} split. 

\subsection{Statistics of the Dataset}
\label{sec:statistics}
We present statistics, automatic measures of coherence and diversity of our dataset to show that our collected stories are more coherent and diverse. 

\myparagraph{Statistics} 
We compare the properties of our dataset to similar previous datasets including Travel blogs \citep{park2015disneyNYC} and VIST \citep{Huang2016} in Table \ref{tab:dataset-statistics}. 
Our VWP dataset has 1965 image sequences with 20763 unique images from 122 movies. Each image sequence has 5 to 10 images. 
Our stories have 45\% more tokens, 103\% more events, and 285\% more characters per text compared to the VIST dataset. While the Travel blogs dataset has longer stories, it has only one image per story. 

\begin{table*}[t]
\centering
\begin{tabular}{l|l|lllllll}

\textbf{Dataset}	&	\textbf{\#}	&	\textbf{PRD}	&	\textbf{Characters}	&	\textbf{Arguments}	&	\textbf{arg0}	&	\textbf{arg1}	&	\textbf{arg2}	&	\textbf{arg-loc}	\\	\hline
VIST	&	998	&	0.063	&	0.184	&	0.055	&	0.041	&	0.018	&	0.018	&	0.013	\\	
VWP (Ours)	&	1000	&	0.068	&	0.21	&	0.057	&	0.101	&	0.048	&	0.025	&	0.017	\\	\hline
									
\end{tabular}
\caption{\label{tab:jaccard} Average Jaccard similarity between stories of each image sequence. All numbers are higher the better except the first column which is the number of image sequences. }
\end{table*}

\begin{table*}[t]
\centering								
\begin{tabular}{l|lllll|lll}																
\textbf{Dataset}	&	\textbf{Voc}	&	\textbf{Verb}	&	\textbf{Verb :}	&	\textbf{Verb :}	&	\textbf{Diverse}	&	\textbf{unigram}	&	\textbf{bigram}	&	\textbf{trigram}			\\	
	&		&		&	\textbf{Voc \%}	&	\textbf{Tok \%}	&	\textbf{Verb \%}	&		&		&				\\	\hline
VIST	&	12627	&	3447	&	27.3	&	1.2	&	73.6	&	3.39	&	33.48	&	75.22			\\	
VWP (Ours)	&	13637	&	4811	&	35.28	&	1.23	&	79	&	2.71	&	34.87	&	79.10			\\	\hline

\end{tabular}																								
\caption{\label{tab:event-diversity} Comparison of diversity. The first column shows the names of the datasets. The next five columns show event diversity for validation split of VIST versus a comparable sample of VWP. We report measures including the vocabulary size (Voc), unique number of verbs (Verb), verb-vocabulary ratio (Verb : Voc \%), verb-token ratio (Verb : Tok \%) and percentage of diverse verbs (Diverse Verb \%). The last three columns show predicate n-grams diversity for VIST versus VWP dataset. We measure diversity using unique:total ratios of predicate unigram, bigram and trigram. All numbers are the higher the better. }
\end{table*}

\begin{table}[t]								
\centering								
\begin{tabular}{l|llll}								
								
\textbf{Label}	&			\textbf{VWP}	&		&	\textbf{VIST}	&		\\	
	&			\#	&	\%	&	\#	&	\%	\\	\hline
E Grounded	&			164	&	54.9	&	38	&	45.2	\\	
E Inferred	&			134	&	44.8	&	39	&	46.4	\\	
E Hallucianted	&			1	&	0.3	&	7	&	8.3	\\	\hline
A Grounded	&			447	&	62.5	&	105	&	53.6	\\	
A Inferred	&			254	&	35.5	&	64	&	32.7	\\	
A Hallucianted	&			14	&	2.0	&	27	&	13.8	\\	\hline

\end{tabular}								
\caption{\label{tab:visual-groundedness}  Visual Groundedness of stories for VIST versus VWP dataset. We report counts and percentages of each label in each data. E means event and A means argument. } 						
\end{table}

\myparagraph{Coherence} 
\label{coherence}
We first analyze coherence of the stories focusing on the characters and their appearances. According to Centering theory \citep{grosz1995centering}, coherent narratives are typically structured such that salient entities often appear in strong grammatical roles like subject or object. As a result, we apply a model based on this theory, Entity Grid \citep{lapata2005automatic}, to measure the local coherence of our dataset. We apply the generative Entity Grid model implemented in the Cohere toolkit \citep{smith2016cohere} on the VIST and our dataset. We calculate the log-likelihood based on entity transitions as the story coherence. The results in Table \ref{tab:local-coherence} show that our dataset is significantly more coherent compared to the VIST (unpaired $t$-test, $t=-5$, $p$-value $<0.001$). 

To further check whether event elements are semantically related given the same image sequence, we also compute the average Jaccard similarities between event elements of the stories for each image sequence by main characters, predicates (without auxiliary verbs), and arguments in different semantic roles. We identify the main characters in the raw text using coreference clusters \cite{Lee2018HigherorderCR}. To ensure that characters mentioned only once in the story can be detected by the coreference resolution model, we append the stories with one introductory sentence per character. For example, to identify the character \textit{Jack} in Figure \ref{fig:datasets-comp}, we add ``\textit{This is Jack.}'' before the story.  
The Jaccard similarity between story A and B is defined as $ J(A, B) = \frac{A \cap B}{A \cup B}$, where $A$, $B$ are the token sets of predicate/argument in story A and B. 
The results in Table \ref{tab:jaccard} show that the event elements of stories conditioned on the same image sequence are more semantically related to each other. Our dataset has higher semantic cohesion compared to VIST dataset.

\myparagraph{Diversity} We then measure diversity of the stories from two perspectives: 1) If a story has a plot with a series of different events, it must have diverse events instead of just repeating one event; 2) If these events are combined into different n-grams in the plot, then the story must have diverse predicate n-grams. For example, in the last column in Figure \ref{fig:datasets-comp}, the character \textit{Will} has a predicate trigram \textit{(tell, convince, work)}, which is different from the next trigram \textit{(convince, work, call)}. 

For event diversity, we follow \citet{fan-etal-2019-strategies} and \citet{goldfarb2020content} to obtain the unique number of verbs, the verb-vocabulary ratio, verb-token ratio and the percentage of diverse verbs (not in the top 5 most frequent verbs). The results in Table \ref{tab:event-diversity} show that our dataset has higher event diversity than VIST  across all measures. 
To measure predicate n-gram diversity, we extract and lemmatize verbs obtained from a Semantic Role Labeling model \cite{shi2019simple} and calculate the unique:total ratios of predicate unigram,  bigram, and trigram (Table \ref{tab:event-diversity}). We observe that the event sequences in VWP are more diverse than VIST, because VWP has a lower unigram ratio but higher bigram and trigram ratios. 

\myparagraph{Visual Groundedness} 
To check the visual groundedness of the stories, we first apply a semantic role labeller to 25 human-written stories each from VWP and VIST. We obtain 299 events and 715 arguments from the VWP samples, 84 events and 196 arguments from the VIST samples. We then manually annotated these events and arguments with 3 labels: 1) \textit{Grounded} means the event or argument is in the corresponding image; 2) \textit{Inferred} means not in the image, but can be inferred; 3) \textit{Hallucianted} means not in image and cannot be inferred. 

The results in Table \ref{tab:visual-groundedness} show that about 55\% of the events and 63\% of the arguments in VWP stories appear in images which are higher than the 45\% of the events and 54\% of the arguments in VIST stories that appear in images. About 45\% of the events and 35\% of the arguments in VWP stories are not in the images but can be inferred from the other images or the previous part of the stories, which are very similar to the results of VIST stories (46\% of the events and 33\% of the arguments not in images but can be inferred). Only 2\% of the arguments in VWP stories are not in the images and cannot be inferred (i.e. not visually grounded). However, there are 8\% of the events and 14\% of arguments are not visually grounded in VIST.


\section{Experiment and Evaluation}

\begin{figure*}[t]
    \centering
    \subfigure[Character grid with high coherence. ]{
        \centering
        \includegraphics[width=0.44\textwidth,trim={0cm 12cm 22cm 0cm},clip]{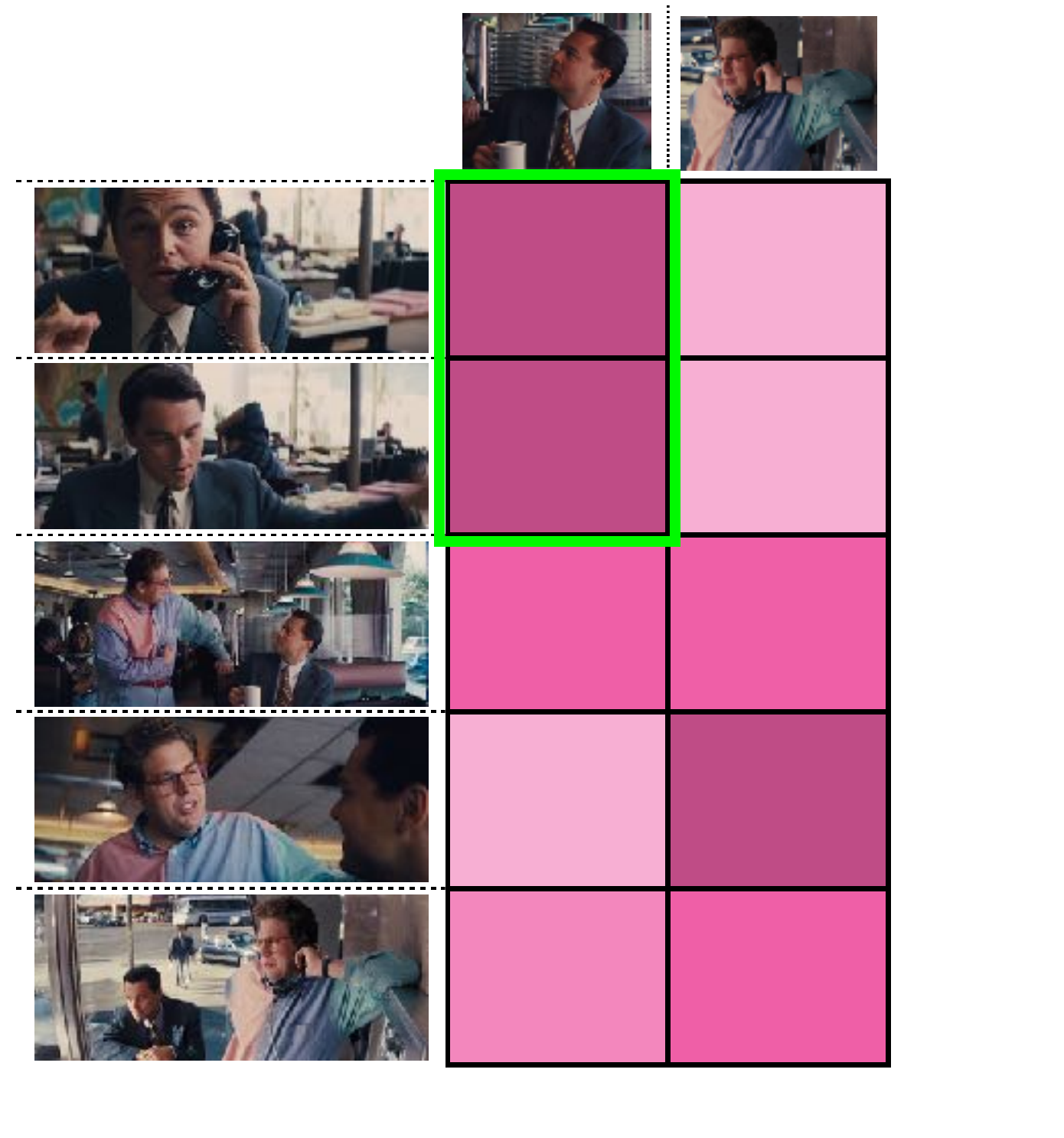}
    }
    \subfigure[Character grid with low coherence. ]{
        \centering
        \includegraphics[width=0.47\textwidth,trim={0cm 8cm 17cm 0cm},clip]{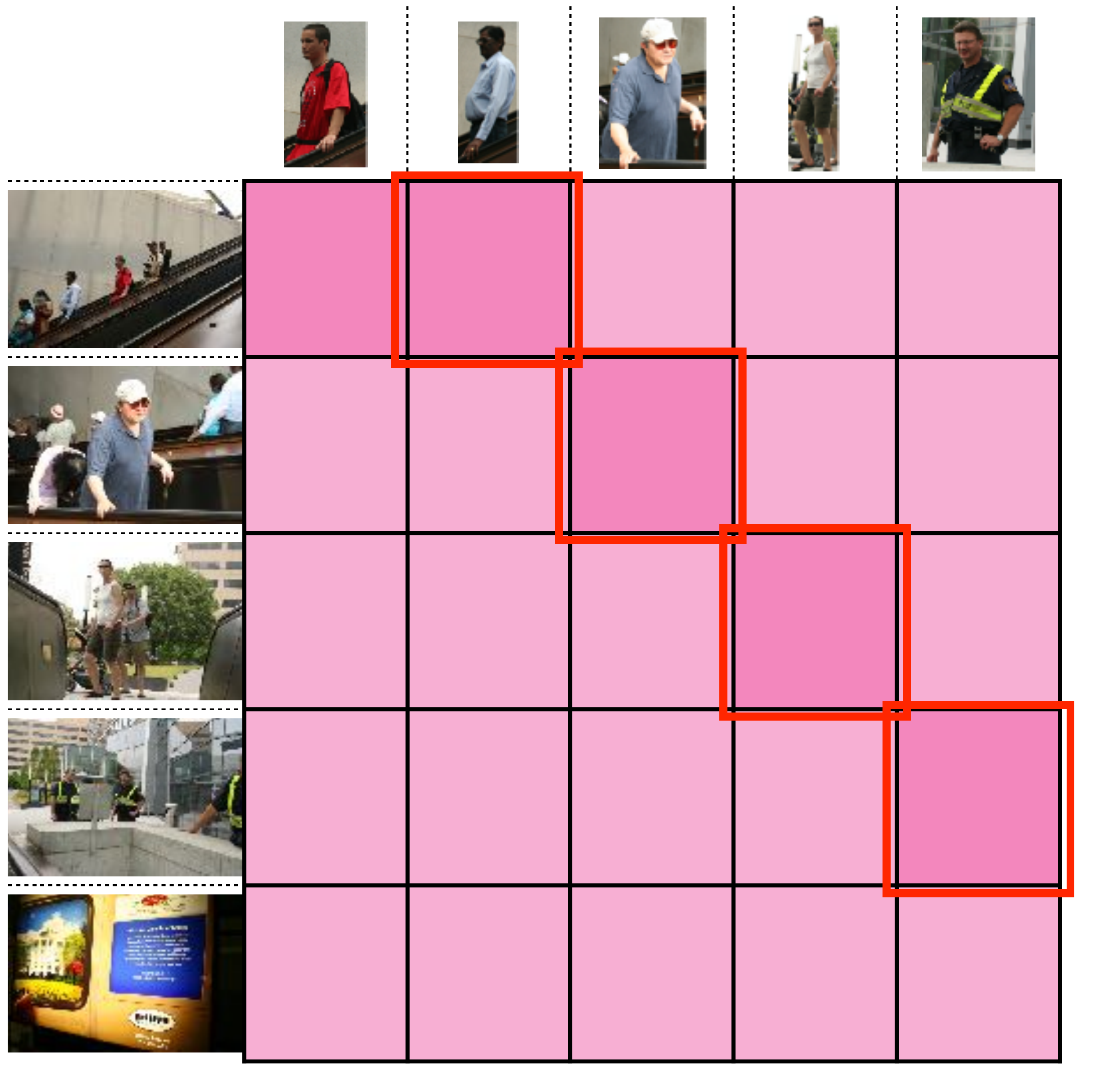}
    }
    \caption{Example of character grid representations. Each row represents an image and each column represents a character. Shades of the cells indicate the similarities between the character features and the image features. The darker colour represents higher similarity. The green square shows the pattern that indicates high coherence and the red square shows the pattern that indicates low coherence. }
    \label{fig:CharGrid}
\end{figure*}

In this section, we propose a strong baseline and show the experimental results on our VWP dataset. Our goal is to demonstrate the usefulness of our dataset. 

We extract features for all images with Swin Transformer \citep{liu2021swin}, a state-of-the-art computer vision backbone model where all parameters are fixed. We use their official model checkpoint, pre-trained on the ImageNet-21K dataset, to increase domain generality. We extract three different visual features: 

\myparagraph{1} Global features (global) are most commonly used in image-based language generation. We extract global features from the output of the last feedforward layer. 

\myparagraph{2} Object features (obj) are widly used in image-based language generation. Since \textit{person} is also a label in object detection \citep{lin2014microsoft}, using object features is a proper baseline for character features. We obtain object features using a Cascade Mask R-CNN object detector \citep{cai2019cascade} with the same Swin Transformer backbone. We crop the bounding boxes of the top 20 objects that the detector predicts for each image and extract the features the same way as global features. 

\myparagraph{3} Character features (char) are extracted by cropping out the least blurry instance of each character using bounding boxes from our dataset. We feed the bounding boxes to the same Swin Transformer backbone and get the features from the last feedforward layer.

We use the following models for visual story generation as baselines: 



\myparagraph{GPT-2} \citep[GPT-2;][]{radford2019language} is a Transformer-based language model pre-trained on large-scale text. We use the small version which is widely used in previous work of story generation. 

\myparagraph{TAPM} \citep[TAPM;][]{yu2021transitional} is a Transformer-based model which adapts the visual features with pre-trained GPT-2. This is the current state-of-the-art model for visual story generation. 

For each baseline, we consider four different variants with different input: 1) only global image features; 2) global features and object features; 3) global features and character features; 4) all three available features.

\begin{figure*}[t]
\includegraphics[width=0.96\textwidth,trim={0cm 17cm 0cm 0cm},clip]{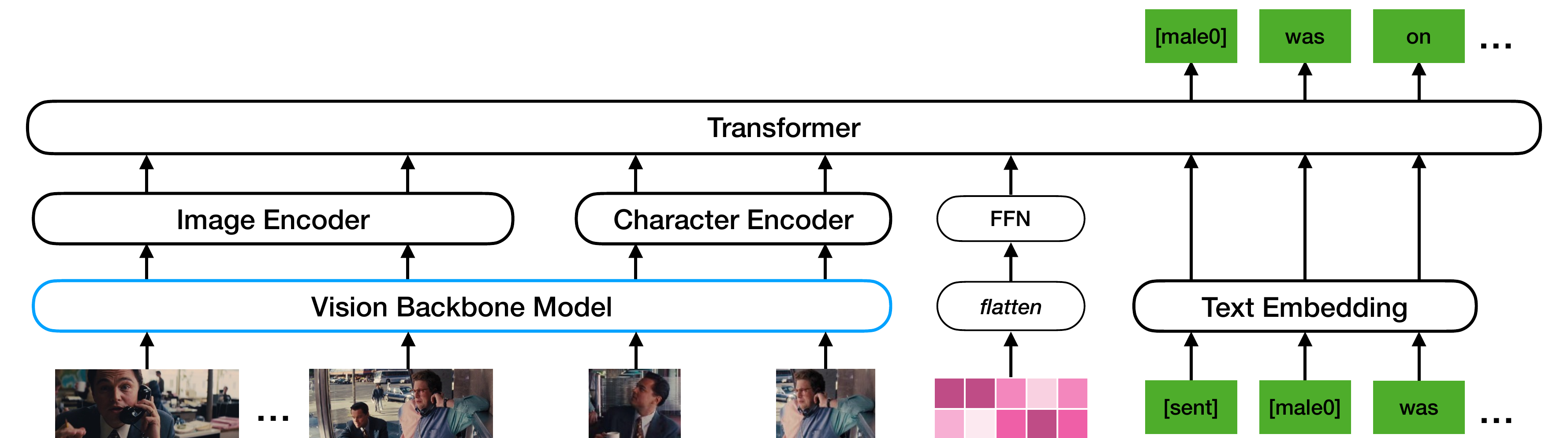}
\caption{ Architecture of character-grid transformer. The blue circles are pre-trained components where the parameters are fixed. }
\label{fig:model}
\end{figure*}

\subsection{Character-based visual story generation}
We propose the character-grid transformer model (CharGrid) as a strong baseline to show the importance of modeling coherence and visual groundedness. We hypothesize that characters and different instances of them in image sequences play an important role in visual story generation models in two dimensions: firstly, explicit character representations can improve groundedness of generated stories, which has been observed in textual stories \cite{clark2018neural}. Secondly, representations that describe different instances of characters across images are beneficial to image-based story generation models.

\myparagraph{Character Grid}
To represent coherence of image sequences, we proposed a novel visual representation, \textit{character grid}. As we mentioned in Section \ref{coherence}, one of the most effective methods to measure text coherence is Entity Grid, a matrix of sentences by entities where the cells are the grammatical roles of the entities in the sentence context  \citep{lapata2005automatic}. The contribution of an entity's mention to the coherence of a sentence is defined by its within-sentence grammatical role. 

Inspired by this, we measure the narrative importance of a character in an image by the similarity between global image features and the character's features. 
We thus measure the coherence of an image sequence using a matrix $\mathbf{C}$ of images by character instances shown in Figure \ref{fig:CharGrid}. We obtain the narrative importance of each character instance by computing the dot product of each character's features and the corresponding global image features. In the character grid $\mathbf{C}$, each element is computed as $\mathbf{c}_{ab} = \mathbf{i}_a\cdot\mathbf{l}_b$, where $\mathbf{i}_a$ is the global features of image $a$, and $\mathbf{l}_b$ is the features of character $b$.

\myparagraph{Model Architecture}
As we show in Figure \ref{fig:model}, the architecture is based on the Transformer model. The input to the Transformer is a sequence of tokenized features including: global images features, character features, character grid, and text features. Global images features and character features are the same as the features for baseline models described above, which are first fed to trainable global and character encoders that consists of a feedforward layer. Text features are tokenized representations of the generated context, which are presented to the model incrementally. The character grid is flattened and fed to a feedforward layer. The four inputs then pass through the transformer module. The output obtained at each time step is a probability distribution over all possible output tokens from a pre-trained GPT-2 tokenizer \citep{wolf2020transformers}. 

We also construct two variants of our model to inspect the contributions of each design decision. We replace the character features with object features to obtain the object-grid transformer model (ObjGrid). We use both character features and object features to obtain the entity-grid transformer model (EntiGrid). 

\myparagraph{Model Training}
We randomly initialized the model parameters except for the vision backbone model. We optimize the model by maximizing the likelihood of the image sequence-story pairs in the training set. The parameters are updated via back propagation. We employ Nucleus sampling \citep{holtzman2019curious} to obtain the full output sequence for validation. We compute the METEOR score \citep{banerjee-lavie-2005-meteor} on the validation set after each training epoch. If the current epoch gets a lower METEOR score, we consider the current epoch as the best epoch and generate stories and run automatic metrics on the test set. We choose the METEOR score following previous work in visual story generation (see Section \ref{sec:vst}). In addition, \citet{Huang2016} found METEOR correlates better with human judgement than BLEU and Skip-Thoughts similarity on the VIST dataset.

\begin{table*}[t]
\centering
\resizebox{\textwidth}{!}{%
\begin{tabular}{lllllllll}

\textbf{Model}	&	\textbf{Features}	&	\textbf{B-1}	&	\textbf{B-2}	&	\textbf{B-3}	&	\textbf{B-4}	&	\textbf{M}	&	\textbf{R-L}	&	\textbf{C}	\\	\hline
GPT-2	&	global	&	38.65**	&	20.28**	&	9.78**	&	4.68*	&	31.64**	&	24.24+	&	1.66**	\\	
GPT-2 + obj	&	global, obj	&	40.65**	&	21.35**	&	10.2**	&	4.87*	&	31.69**	&	24.05+	&	1.85**	\\	
GPT-2 + char	&	global, char	&	39.95**	&	21.04**	&	10.11**	&	4.92+	&	31.85*	&	24.19+	&	1.57**	\\	
GPT-2 + obj,char	&	global, obj, char	&	40.41**	&	21.44**	&	10.56**	&	5.06+	&	32.03*	&	24.38	&	1.87**	\\	
TAPM	&	global	&	39.85**	&	21.7**	&	10.72**	&	5.19	&	32.38+	&	25.09	&	1.48**	\\	
TAPM + obj	&	global, obj	&	40.86**	&	22.13**	&	10.83**	&	5.25	&	32.34+	&	24.91	&	1.82**	\\	
TAPM + char	&	global, char	&	40.03**	&	21.68**	&	10.66**	&	5.18	&	32.42+	&	24.88	&	1.4**	\\	
TAPM + obj,char	&	global, obj, char	&	40.87**	&	21.99**	&	10.72**	&	5.06+	&	32.48+	&	24.87	&	1.59**	\\	\hline
\textit{Ours}	&		&		&		&		&		&		&		&		\\	\hline
ObjGrid	&	global, obj	&	47.66	&	25.26	&	11.95	&	5.42	&	32.83	&	24.42	&	4.68	\\	
EntityGrid	&	global, obj, char	&	45.83	&	24.85	&	\textbf{12.11}	&	\textbf{5.7}	&	32.68	&	24.89	&	3.53+	\\	\hline
CharGrid 	&	global, char	&	\textbf{47.71}	&	\textbf{25.33}	&	11.95	&	5.42	&	\textbf{33.03}	&	\textbf{25.01}	&	\textbf{4.83}	\\	\hline

\end{tabular}
}

\caption{\label{tab:eval-reference} Results of all models using different input features on the test set of VWP using reference-based metrics including BLEU (B), METEOR (M), ROUGE-L (R-L), and CIDEr (C). All numbers are average of three runs with different random seeds. (pretrain) indicates models initialised the Transformer with GPT-2 pre-trained weights. +, * and ** represent that the number is one, two or three standard deviations away from the mean of CharGrid model. }

\end{table*}

\subsection{Reference-based metrics} 
Our goal is to show the effectiveness of character grid representations. 
Although it has been shown that reference-based metrics correlate poorly with human judgements in open-ended language generation tasks \citep{guan2020union,gehrmann-etal-2021-gem}, it is still efficient to use them for comparisoon across many different models. Furthermore, we want to make our results comparable to the original results of the state-of-the-art model TAPM \citep{yu2021transitional}. They applied greedy search to generate stories with their models for testing and reported reference-based metrics. We thus follow the same setting and compare our proposed CharGrid model against several previous baselines. 

We train all the models for at most 15 epochs with 3 different random seeds. We apply the reference-based metrics including unigram (B-1), bigram (B-2), trigram (B-3), and 4-gram (B-4) BLEU scores~\citep[B;][]{papineni-etal-2002-bleu}, METEOR~\citep[M;][]{banerjee-lavie-2005-meteor}, ROUGE-L~\citep[R;][]{lin-2004-rouge}, and CIDEr~\citep[C;][]{vedantam2015cider}, which were used in the visual storytelling shared task \cite{ws-2018-storytelling}. We then report the mean and standard deviation of 3 runs. 

Results in Table \ref{tab:eval-reference} show that the character-grid transformer model (CharGrid) driven by visual coherence outperforms TAPM with character features (TAPM + char) significantly on BLEU-1/2/3 and CIDEr and marginally on METEOR. CharGrid model also outperforms GPT-2 with character features (GPT-2 + char) significantly on most metrics except marginally on BLEU-4 and METEOR. 
The object-grid transformer model (ObjGrid) outperforms TAPM with object features (TAPM + obj) significantly on BLEU-1/2/3 and CIDEr and marginally on METEOR. ObjGrid model also outperforms GPT-2 with object features (GPT-2 + obj) significantly on most metrics except marginally on BLEU-4. 
The entity-grid transformer model (EntiGrid) outperforms TAPM with all features (TAPM + obj,char)significantly on most metrics except marginally on METEOR and  ROUGE-L. EntiGrid model also outperforms GPT-2 with all features (GPT-2 + obj,char) on most metrics except BLEU-4. 
These results show the effectiveness of character/object/entity grid representations for coherence of image sequences.



\begin{table*}[t]
\centering
\begin{tabular}{l|lllll}

\textbf{Model}			&	\textbf{Grammatical}	&	\textbf{Coherence}	&	\textbf{Visual Groundedness}	&	\textbf{Diversity}							\\	\hline
TAPM + char vs. TAPM           &   +2.45    &   +1.99    &   +3.99*   &   +1.69                            \\  \hline
CharGrid vs. TAPM + char            &   +6.49**  &   +8.41**  &   +6.25*   &   +11.06**                         \\  \hline

\end{tabular}
\caption{\label{tab:eval-model-human} Human binary judgments in percentage of generated stories between TAPM and TAPM with character features (TAPM + char), TAPM + char and our model (CharGrid) on test set of VWP across four criteria: \textit{Grammaticality, Coherence, Visually Groundedness} and \textit{Diversity}. The numbers are percentages. * means $p$-value $<0.05$. ** means $p$-value $<0.01$. }
\end{table*}

\begin{figure*}[t]
\includegraphics[width=0.95\textwidth,trim={0cm 15.5cm 0cm 0cm},clip]{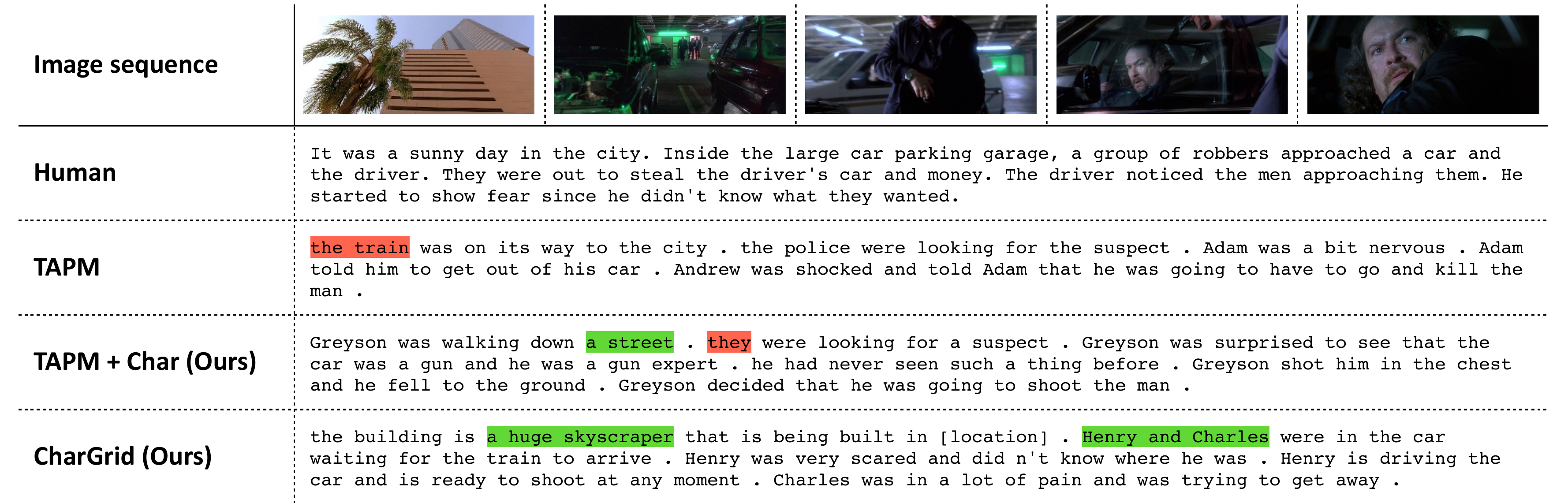}
\caption{ Qualitative results of generated and human-written stories. Red colour represents errors made by models and green colour indicates better output. }
\label{fig:output}
\end{figure*}

\subsection{Human evaluation} 
Because story generation is an open-domain task, reference-based metrics can only show how output stories match with the references. In order to measure the quality of generated stories directly, we conduct a crowdsourcing experiment to obtain human binary judgments between two systems. 
We design the first question for \textit{Grammaticality}, which measures whether the textual outputs are at least grammatical and sets a foundation for other metrics. We then design  questions for two properties that we identified for good textual stories: \textit{Coherence} and \textit{Diversity}.  Finally, we ask a question to compare the \textit{Visual Groundedness} in order to make sure that the stories are relevant to the input image sequence. 

We conduct the experiment with 28 crowd workers over 50 pairs of stories and report the percentage of the judgments for each system that annotators are in favor of. To make the stories more readable, we change the generated character placeholders to randomly sampled names. The results in table \ref{tab:eval-model-human} show that TAPM with character features (TAPM + char) outperforms TAPM in Visual Groundedness significantly. CharGrid outperforms TAPM + char on all metrics significantly. We use two-sided binomial tests. This indicates that our character grid representation yields better stories. These results confirm the findings in the evaluation with reference-based metrics.

\subsection{Qualitative evaluation}
We also conduct a qualitative evaluation to show that stories generated by TAPM with character features are more visually grounded than without character features and character grid representation further improves the coherence and visual groundedness. To obtain more diverse text, we use Nucleus Sampling \citep{holtzman2019curious} with $p=0.1$ on all models to generate the stories. As in Figure \ref{fig:output}, TAPM generates unreasonable noun phrases \textit{the train}. With character features, TAPM + char is able to explore character-object interaction and reason that there is no train in the image. So it generates more reasonable terms \textit{a street}. 

However, TAPM + char model fails to represent the relations between characters, TAPM + char generates pronoun \textit{they} without introducing characters in the second image. In contrast, CharGrid introduces two new characters correctly.


\section{Conclusions and Future Work}
We show that curated image sequences with characters are effective as writing prompts for visual story generation in both data collection and model design. By filtering images without any objects and removing highly similar images to boost diversity, we can improve visual tellability of image sequences. 
Presenting selected characters during the story-writing yields stories with characters grounded in images, which are more coherent and have more narrativity. Correspondingly, using character features as input to the story generation model can improve the quality of generated stories. Adding the character grid representation can bring further improvements in coherence, grammaticality, visual groundedness, and narrativity. 

\myparagraph{Future work} 
One important property of visual narratives not covered in this work is \textit{narrativity} \citep{piper-etal-2021-narrative}, i.e.~whether an image sequence contains necessary narrative structures to be a narrative. A narrative structure can be achieved by events following a typical order with roles like \textit{Establisher}, \textit{Initial}, \textit{Initial
Peak} and \textit{Release} \citep{cohn2013visual}. We observe that these roles of events emerge in our collected stories. Our annotation of different instances of the same character across a story allow us to construct event chains for each character. Future work should investigate how to annotate roles of these events, measure narrativity and build a model to generate stories with higher narrativity. 

A major assumptions of all previous work in storytelling is that all humans are equally and reasonably proficient in story-writing and can translate visual narratives into textual narratives. However, individual differences in the writing proficiency of humans must have an impact on story quality. How to explore this from the perspective of both data selection and model design would be an interesting future direction to take. 



\bibliographystyle{acl_natbib}
\bibliography{egbib}

\end{document}